\documentclass[sn-basic,square,comma,numbers,sort&compress]{sn-jnl}
\setcitestyle{numbers}

\makeatletter
\renewcommand\@biblabel[1]{#1.}
\makeatother
\usepackage{array}
\newcolumntype{P}[1]{>{\centering\arraybackslash}p{#1}}

\usepackage[export]{adjustbox}
\usepackage{tablefootnote}
\usepackage{lineno,hyperref}
\usepackage{dirtytalk}
\usepackage{color}
\usepackage{mathtools}
\usepackage{comment}
\usepackage{rotating}
\usepackage{cancel}
\usepackage{algorithm}
\usepackage{algpseudocode}
\usepackage{multirow}
\usepackage{multicol}
\usepackage{graphicx}
\usepackage{booktabs}
\usepackage{listings}
\usepackage[multiple]{footmisc}
\usepackage[skip=0.2\baselineskip]{caption}
\usepackage{mathtools}
\usepackage{comment}
\usepackage[T1]{fontenc}
\usepackage{soul,color}
\usepackage{pifont}
\newcommand{\cmark}{\ding{51}}%
\newcommand{\xmark}{\ding{55}}
\jyear{2022}
\usepackage{array}
\newcolumntype{P}[1]{>{\centering\arraybackslash}p{#1}}
\modulolinenumbers[5]

\begin{document}

\title[Automatic Detection of Relevant Information, Predictions and Forecasts]{Automatic Detection of Relevant Information, Predictions and Forecasts in Financial News through Topic Modelling with Latent Dirichlet Allocation}

\author*[1]{\fnm{Silvia} \sur{García-Méndez}}\email{sgarcia@gti.uvigo.es}
\equalcont{These authors contributed equally to this work.}

\author[1]{\fnm{Francisco} \sur{de Arriba-Pérez}}\email{farriba@gti.uvigo.es}
\equalcont{These authors contributed equally to this work.}

\author[1]{\fnm{Ana} \sur{Barros-Vila}}\email{abarros@gti.uvigo.es}
\equalcont{These authors contributed equally to this work.}

\author[1]{\fnm{Francisco J.} \sur{González-Castaño}}\email{javier@gti.uvigo.es}
\equalcont{These authors contributed equally to this work.}

\author[1]{\fnm{Enrique} \sur{Costa-Montenegro}}\email{kike@gti.uvigo.es}
\equalcont{These authors contributed equally to this work.}

\affil[1]{\orgname{Information Technologies Group, atlanTTic, University of Vigo}, \orgaddress{\street{E.I. Telecomunicaci\'on, Campus Lagoas-Marcosende}, \city{Vigo}, \postcode{36310}, \country{Spain}}}

\abstract{Financial news items are unstructured sources of information that can be mined to extract knowledge for market screening applications. They are typically written by market experts who describe stock market events within the context of social, economic and political change. Manual extraction of relevant information from the continuous stream of finance-related news is cumbersome and beyond the skills of many investors, who, at most, can follow a few sources and authors. Accordingly, we focus on the analysis of financial news to identify relevant text and, within that text, forecasts and predictions. We propose a novel Natural Language Processing (\textsc{nlp}) system to assist investors in the detection of relevant financial events in unstructured textual sources by considering both relevance and temporality at the discursive level. Firstly, we segment the text to group together closely related text. Secondly, we apply co-reference resolution to discover internal dependencies within segments. Finally, we perform relevant topic modelling with Latent Dirichlet Allocation (\textsc{lda}) to separate relevant from less relevant text and then analyse the relevant text using a Machine Learning-oriented temporal approach to identify predictions and speculative statements. Our solution outperformed a rule-based baseline system. We created an experimental data set composed of 2,158 financial news items that were manually labelled by \textsc{nlp} researchers to evaluate our solution. Inter-agreement Alpha-reliability and accuracy values, and \textsc{rouge-l} results endorse its potential as a valuable tool for busy investors. The \textsc{rouge-l} values for the identification of relevant text and predictions/forecasts were 0.662 and 0.982, respectively. To our knowledge, this is the first work to jointly consider relevance and temporality at the discursive level. It contributes to the transfer of human associative discourse capabilities to expert systems through the combination of multi-paragraph topic segmentation and co-reference resolution to separate author expression patterns, topic modelling with \textsc{lda} to detect relevant text, and discursive temporality analysis to identify forecasts and predictions within this text. Our solution may have compelling applications in the financial field, including the possibility of extracting relevant statements on investment strategies to analyse authors' reputations.}


\maketitle

\section{Introduction}

\subsection{Motivation}

New efficient algorithms \cite{Manogaran2018,Delic2019,Ma2020,Zuo2020} and the prolific sources of online information have boosted applied data analysis research. In this scenario, Natural Language Processing (\textsc{nlp}) techniques are being successfully applied to unstructured textual data \cite{Guetterman2018,Zhang2019,Balyan2020,Lu2022}, from the simplest approaches that use morphological information as input \cite{Kumar2019} to more complex methodologies that take advantage of syntactic patterns and semantic relations \cite{K.2018}. 

Financial Knowledge Extraction (\textsc{ke}) is of particular interest. \textsc{nlp} techniques have been used to apply a wealth of market forecasting research to financial news, economic reports and financial expert comments \cite{Xing2018}. Financial news describes relevant market events, their causes and their possible effects. Transferring human associative discourse capabilities \cite{Lytos2019} from this type of content is challenging.

\subsection{Financial knowledge extraction}

Some representative examples of financial \textsc{ke} include information extraction from financial news for firm-based monitoring \cite{Kelly2018}; analysis of financial risk such as volatility \cite{Atkins2018} and Personal Finance Management applications \cite{Isa2018}, among other interesting use cases \cite{Balyan2020}. Most of these \textsc{ke} systems engineer specific features of the content with their target in mind \cite{Zhang2019Narratives}. 

It is well known that there is a strong relation between mass media news and stock market state \cite{Cepoi2020,Swathi2022}. Previous research has shown that information published in media outlets or shared financial data in printed media, radio, television, and web sites is correlated with future stock market events \cite{LOUGHRAN2016}. Apart from providing valuable objective information in financial news, authors speculate about market events within political, social and cultural contexts. In these unstructured texts the discourse flows around certain key statements and predictions, and an automatic financial news analysis system should distinguish between less relevant data and predictions to gather knowledge to assist investors in decision making \cite{Lutz2020}.

\subsection{Temporality at the discursive level}

Temporal representation in texts and speculative statements in particular is based on semantic combinations of certain linguistic structures and elements \cite{Mohamed2019}. However, the vast majority of works on temporality research at the discursive level have simply focused on verb tenses \cite{Evers-Vermeul2017}, ignoring their semantic context.

\subsection{Research goal and main contribution}

Our research is a case of financial News Analysis (\textsc{na}) \cite{Kelly2018,Atkins2018,Isa2018} within the field of Intelligence Amplification, which has lately gained attention \cite{Jang2019,Chau2019,Phi2020} as a means of enhancing the understanding and reasoning capabilities of automatic \textsc{ke} solutions and transferring human associative discourse capabilities to expert systems. Our case contributes to solving the problem of extracting relevant text from financial news and, within that relevant text, identifying forecasts and predictions. Our solution may be valuable in helping inexpert stockholders to process more financial news more efficiently.

To the best of our knowledge, this is the first study to propose an approach for the automatic detection of relevant events in financial \textsc{na} based on the joint consideration of relevance and temporality analysis at the discourse level. 

\subsection{Approach}
\label{contributions}

Our approach comprises:
\begin{itemize}
 \item Multi-paragraph topic segmentation and co-reference resolution to separate author expression patterns.
 \item Detection of relevant text through topic modelling with Latent Dirichlet Allocation (\textsc{lda}), outperforming a rule-based system.
 \item Identification of forecasts and predictions within relevant text using discursive temporality analysis and Machine Learning (\textsc{ml}).
\end{itemize}

We demonstrate the performance of these features using an experimental data set composed of news items from widely used financial sources. The final data set had 2,158 financial news items similar in size to or even larger than other studies in the literature \cite{Li2018,Long2019,Al-Smadi2019,Zhang2020,DeOliveiraCarosia2021,Xie2021} that were manually labelled by \textsc{nlp} researchers.

\subsection{Structure of the paper}
\label{organization}

The rest of this article is organised as follows. Section \ref{sec:related_work} reviews related work on \textsc{ke} and \textsc{na} solutions. Section~\ref{sec:system} describes our automatic system for detecting relevant financial events based on \textsc{nlp} and \textsc{ml} techniques. Section \ref{sec:results} presents the text corpus and numerical evaluation of our solution. Finally, Section \ref{sec:conclusions} concludes the article.

\section{Related work}
\label{sec:related_work}

Stock market research is based on fundamental and technical approaches \cite{Nti2020}. Fundamental approaches involve performing stock market forecasts using numerical data such as price variation. Technical approaches, in turn, focus on the temporal dimension of financial events. They apply trend modelling techniques to historical asset data forecasts.

Previous research on Data Mining and \textsc{ke} for stock market screening on textual data has considered financial news \cite{Atkins2018,Carta2021}, stockholder comments in blogs \cite{Swathi2022} and social networks \cite{Khan2022}. These systems apply \textsc{nlp} techniques \cite{Xing2018} or \textsc{ml} models \cite{Delic2019,Atkins2018}, which may be supervised \cite{Rustam2020}, relying on automatic or manually annotated data sets, or unsupervised \cite{Solorio-Fernandez2020}, taking into account the peculiarities of input data and descriptive patterns. The simplest approach consists of using a vector representation of the content and weighting the terms once meaningless elements, such as prepositions \cite{Garcia-Mendez2020}, are removed. More complex approaches, like the one presented by De Arriba-Pérez \textit{et al.} (2020) \cite{DeArriba-Perez2020}, seek to identify syntactic and semantic patterns as key descriptors of financial news through lexica, grammar and name entity recognition techniques. 

Traditional extraction methods for filtering relevant text in this context comprise manual\footnote{\textit{E.g.}, cascaded and non-deterministic finite state automatons, semantic information extraction solutions, etc.} and automatic\footnote{\textit{E.g.}, supervised and unsupervised \textsc{ml} methods, being the first more extended \cite{Beliga2015}.} pattern discovery approaches \cite{kaiser2005information}. The former require large knowledge bases, such as dictionaries and lexica, and rule sets. They tend to be constrained by specific application domains. Automatic approaches include simple statistical and more demanding, complex linguistic approaches, in addition to the previously mentioned \textsc{ml} solutions \cite{Beliga2015}. \textsc{tf}-\textsc{idf} \cite{Li2018clustering} is remarkably simple, but it has been reported to under-perform on professional texts, as in our case. Alternative solutions combine the previous techniques with knowledge heuristics such as position, length and text format information. A more competitive solution is fuzzy logic for sentence scoring. However, this lacks adaptability and requires manual rule generation, which directly affects performance \cite{Azhari2017}.

Unsupervised extraction is highly practical because it eliminates the burden of text tagging. Nevertheless, many \textsc{ke} solutions rely on supervised methodologies. Examples those of Gottipati S \textit{et al.} (2018) \cite{Gottipati2018}, who designed an \textsc{ml} course improvement solution based on student feedback and compared its performance to a rule-based method; López-Úbeda \textit{et al.} (2021) \cite{Lopez-Ubeda2021}, who extracted relevant information from radiological reports; and Verneer \textit{et al.} (2019) \cite{Vermeer2019}, who proposed a relevance detection system from social media messages (although they noted the great potential of \textsc{lda} as an alternative).

Among extraction solutions developed to detect relevant topics from news pieces (setting aside temporality analysis), Jacobs \textit{et al.} (2018) \cite{Jacobs2018} developed a supervised model for economic event extraction in English news using a sentence-level classification approach, as in our case; Oncharoen \textit{et al.} (2018) \cite{Oncharoen2018} applied the Open Information Extraction system to represent the news data as tuples (actor, action and object); Carta \textit{et al.} (2021) \cite{Carta2021event} employed a real-time domain-specific clustering-based approach for event extraction in news; and Harb \textit{et al.} (2008) \cite{Harb2008} presented a linguistic-based opinion extraction system for blogs.

Assuming there is a direct causal relationship between financial news and asset prices \cite{Cepoi2020}, some authors have explored both \textsc{ml} and other sophisticated techniques such as deep learning to gather context-dependent information for stock market screening \cite{BL2021}. Worthy of note in this respect is the Naive Bayes model by Atkins \textit{et al.} (2018) \cite{Atkins2018} for predicting stock market volatility, which employed as input word-topic correspondence feature vectors obtained with \textsc{lda}. Unlike our proposal, this model considered news content as a whole and did not differentiate non-relevant from relevant parts for their target application. Shilpa \& Shambhavi (2021) \cite{BL2021} presented a prediction framework based on sentiment analysis and stock market technical-indicator features. Temporality was not considered.

Prior work has addressed linguistic \cite{Genc2020}, template-based \cite{Y.2018} and statistical news summarisation approaches \cite{Meena2020}. State-of-the-art summarisation systems may be extractive \cite{Meena2020} or abstractive \cite{Gupta2019}. Extractive summaries, which are more akin to our goal, extract key sentences directly from the input text. These sentences are ranked by importance and selected if they pass a threshold. Query-focused and update summarisation approaches \cite{Alhoshan2020} also deserve consideration as they retrieve information tailored to a specific audience. The summarisation of online financial news in our work focuses on financial investors. The temporal dimension, expressed as discursive temporality, is crucial to us because relevant text in finance-related news may include in addition to factual information, speculations or predictions, whether quantitative or not. In further relation to summarisation, template-based systems on financial \textsc{na} \cite{Barros2019} were limited in early research due to their computational load, the laborious task of defining the templates and their lack of flexibility.

\textsc{ke} solutions, and in particular, financial \textsc{na} systems, have not paid sufficient attention to temporal analysis. The vast majority simply use temporal references provided by timestamps or verb tenses. Evers-Vermeul \textit{et al.} (2017) \cite{Evers-Vermeul2017}, for example, simply noted that as linguistic markers, verb tense suffixes express temporal order and coherence relations through text. Our work goes a step further by analysing sentence-level temporality through syntax and semantics, and detecting temporal elements, expressions and the patterns in which they are arranged.

\begin{table}[!ht]
\small
\caption{Comparison with related works.}~\label{tab:Comparison}
\small
\begin{tabular}{p{2.25cm}P{1.25cm}P{1.5cm}P{1.75cm}p{3cm}} 
\toprule
\multirow{2}{*}{\centering\textbf{Authorship}} & \multirow{2}{*}{\textbf{Source}} & \textbf{Relevance} & \textbf{Temporality} & \multirow{2}{*}{\textbf{Technique}} \\
& & \textbf{analysis} & \textbf{analysis}\\

\midrule

Atkins A \textit{et al.} (2018) \cite{Atkins2018} & \multirow{2}{*}{News} & \multirow{2}{*}{\xmark} & Time series data only & \multirow{2}{*}{\textsc{lda} and \textsc{ml}}\\\\

Oncharoen P \textit{et al.} (2018) \cite{Oncharoen2018} & \multirow{3}{*}{News} & \multirow{3}{*}{\xmark} & Temporal relations only & \multirow{3}{*}{\textsc{cnn} and \textsc{lstm}}\\\\

De Arriba-Pérez F \textit{et al.} (2020) \cite{DeArriba-Perez2020} & Micro-blogging data & \multirow{3}{*}{\xmark} & \multirow{3}{*}{\xmark} & \multirow{3}{*}{\textsc{nlp} and \textsc{ml}}\\\\

Carta S \textit{et al.} (2021) \cite{Carta2021event} & News and micro-blogging data & \multirow{3}{*}{\cmark} & \multirow{3}{*}{\xmark} & \multirow{3}{*}{Clustering}\\\\

Shilpa B \textit{et al.} (2021) \cite{BL2021} & \multirow{2}{*}{News} & \multirow{2}{*}{\xmark} & \multirow{2}{*}{\xmark} & \multirow{2}{*}{\textsc{nlp} and \textsc{ml}}\\

\midrule

\multirow{5}{*}{\textbf{Our proposal}} & \multirow{5}{*}{News} & \multirow{5}{*}{{\large \cmark}} & 
\multirow{5}{*}{{\large \cmark}} & Multi-paragraph topic segmentation, co-reference, \textsc{lda}, discursive analysis and \textsc{ml}\\
\bottomrule
\end{tabular}
\end{table}

Summing up, Table \ref{tab:Comparison} compares the most relevant work related to our proposal.
Our main contribution is the detection of relevant statements on financial news including forecasts and predictions. To do this, our system automatically groups related data and filters out background information. In brief, we present a novel technique combining \textsc{lda} analysis of automatically segmented news with temporality analysis at the discourse level. To our knowledge, this is the first \textsc{na} approach that jointly considers relevance and temporality at the discourse level.

\section{System architecture}
\label{sec:system}

\begin{figure*}[!htbp]
\centering
\includegraphics[scale=0.13]{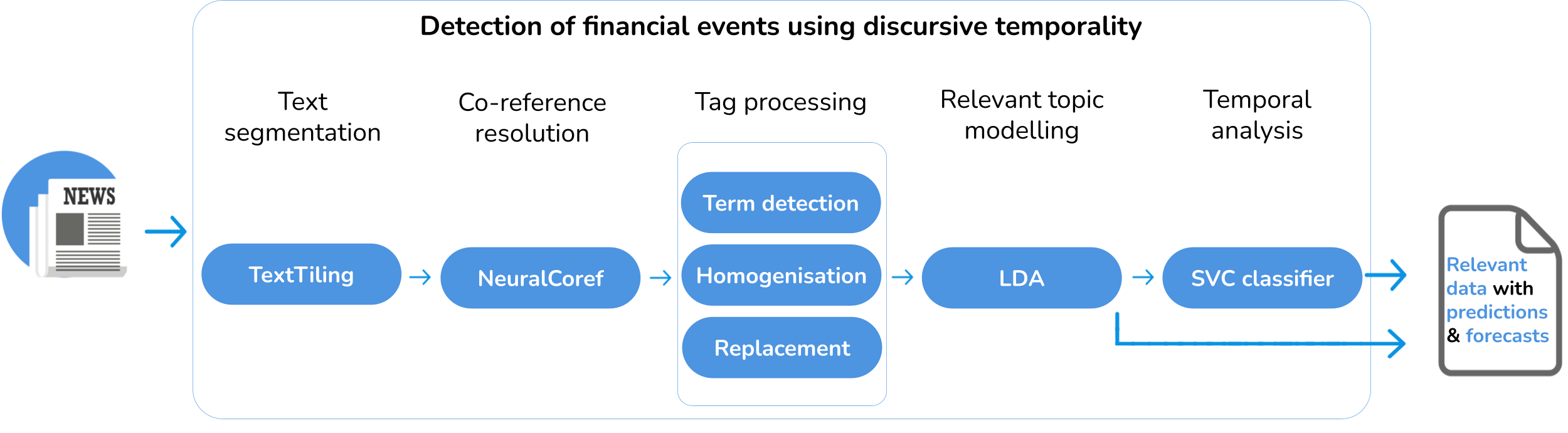}
\caption{\label{architecture} Proposed system.}
\end{figure*}

In this section, we describe our system for the automatic detection of relevant financial events using \textsc{nlp} techniques and \textsc{ke} algorithms. Figure \ref{architecture} shows the scheme of the system. First, we segment the input text to group together closely related information. Then, we apply co-reference resolution to discover internal dependencies in the news content among key references to assets. The next stage is the tag processing stage, which consists of the detection, homogenisation and replacement of financial terms. This is followed by relevant topic modelling and temporal analysis to identify predictions and speculative statements, and ultimately provide investors with a synthesised version of financial news highlighting pertinent information, such as asset performance and forecasts/predictions as a summary.

\subsection{Multi-paragraph topic segmentation}
\label{textiling}

The first stage of the multi-paragraph topic segmentation process applies the enhanced version of the TextTiling algorithm \cite{He2020} to segment news content into subtopic paragraphs.

TextTiling exploits lexical co-occurrence and discourse distribution patterns to identify subtopic paragraphs within a text with reasonable precision from a human's perspective. The algorithm compares, in sequence, the similarity of adjacent text divisions of similar length. If the vocabulary in the first and second parts of the comparison differ, the division is considered a split point. We set a minimum text length of 500 characters to apply the algorithm. Otherwise, no segmentation is performed.

The rationale behind this first stage is the assumption that text segments must be coherent, self-contained information units. Explicit paragraphs of financial news are less useful in our case because, quite often, they just break up the text layout to facilitate readability. It is the informal, inner structure based on key statements and predictions about certain assets or stock markets that interests us.

Table \ref{news} shows an example of the original content of a financial news item. Table \ref{news_texttiling} shows the same item after segmentation with TextTiling (for conciseness, we present just the first two segments). In this example, the first paragraph refers to the current state of the asset and the second describes its past performance. Logically, TextTiling does not always guarantee such a degree of coherence, but in our case, it contributes to the overall efficiency of our system as a building block.

\begin{table*}[!htbp]
\centering
\caption{\label{news} Example of news entry.}
\begin{tabular}{c}
\toprule
\begin{tabular}[c]{@{}p{11cm}@{}}
Verizon Communications (\textsc{nyse:vz}) is proving to be a stable but undervalued company in the market today. In fact, \textsc{vz} stock is worth at least 55\% more than its price today using an analysis of its dividend yield, its own P/E ratio history, and a comparison with its peers. The company reported “boring” earnings, according to Barron’s magazine for Q2 on July 24. But the magazine says “boring is good” in this market. They point out that the communications business is not a bad place to be in a pandemic. For example, Verizon said that its earnings on a non-GAAP adjusted basis was \$1.18 per share. This was only 3\% lower than a year ago. On an adjusted \textsc{ebitda} (earnings before interest, taxes, depreciation, and amortization) basis, its cash flow was down 5\%. However, Verizon generated higher cash flow. For example, first-half 2020 cash flow from operations of \$23.6 billion, an increase of \$7.7 billion from first-half of 2019. This represents a huge increase of over 206\%. Moreover, its free cash flow (\textsc{fcf}) in the first half was \$13.7 billion, an increase of 74.1 percent year over year. But consider this. Verizon trades for a paltry 12.3 times this year’s expected earnings and just 12 times next year[...] 
\end{tabular}\\\bottomrule
\end{tabular}
\end{table*}

\begin{table*}[!htbp]
\centering
\caption{\label{news_texttiling} Example of news entry after applying multi-paragraph TextTiling segmentation.}
\begin{tabular}{cc}
\toprule
\textbf{ID} & \textbf{TextTiling segment} \\\hline
0 & 
\begin{tabular}[c]{@{}p{10cm}@{}}
Verizon Communications (\textsc{nyse:vz}) is proving to be a stable but undervalued company in the market today. In fact, \textsc{vz} stock is worth at least 55\% more than its price today using an analysis of its dividend yield, its own P/E ratio history, and a comparison with its peers.
\end{tabular}\\\hline
1 & 
\begin{tabular}[c]{@{}p{10cm}@{}}
The company reported “boring” earnings, according to Barron’s magazine for Q2 on July 24. But the magazine says “boring is good” in this market. They point out that the communications business is not a bad place to be in a pandemic. For example, Verizon said that its earnings on a non-GAAP adjusted basis was \$1.18 per share[...]
\end{tabular}
\\\bottomrule
\end{tabular}
\end{table*}

\subsection{Co-reference resolution}
\label{section:coref}

The purpose of co-reference resolution is to replace references with meaningful words, which improves the performance of the subsequent \textsc{lda} stage. Specifically, after segmenting the text, we use the Neural Network (\textsc{nn}) by Clark \textit{et al.} (2016) \cite{Clark2016} to generate high-dimensional vector representations for co-reference compatibility of cluster pairs. The \textsc{nn} is composed of two task-oriented sub-networks: a mention-pair encoder and a cluster-pair encoder, which create the distributed representations, and the cluster and mention-ranking models to score the pairs of clusters.

Table \ref{news_coreference} provides an example of a news entry after this procedure. Note how implicit asset references have been replaced by explicit ones, which is essential for the next stages. 

\begin{table*}[!htbp]
\centering
\caption{\label{news_coreference} Example of news entry before and after applying co-reference resolution.}
\begin{tabular}{cc}
\toprule
\textbf{Before co-reference resolution} & \textbf{After co-reference resolution} \\\hline
\begin{tabular}[c]{@{}p{5.5cm}@{}}
Verizon Communications (\textsc{nyse:vz}) is proving to be a stable but undervalued company in the market today. In fact, \textsc{vz} stock is worth at least 55\% more than \textbf{its} price today using an analysis of \textbf{its} dividend yield, \textbf{its} own P/E ratio history, and a comparison with \textbf{its} peers.
\end{tabular} & 
\begin{tabular}[c]{@{}p{5.5cm}@{}}
Verizon Communications (\textsc{nyse:vz}) is proving to be a stable but undervalued company in the market today. In fact, \textsc{vz} stock is worth at least 55\% more than \textbf{\textsc{vz} stock} price today using an analysis of \textbf{\textsc{vz} stock} dividend yield, \textbf{\textsc{vz} stock} own P/E ratio history, and a comparison with \textbf{\textsc{vz} stock} peers.
\end{tabular}
\\\bottomrule
\end{tabular}
\end{table*}

\subsection{Tag processing: financial term detection, homogenisation and replacement}
\label{processing}

After text segmentation and co-reference resolution, the tag processing stage homogenises the input for the subsequent \textsc{lda} stage. First, asset identifiers are detected using our financial lexica\footnote{\label{gtiresources}Available at \url{https://www.gti.uvigo.es/index.php/en/resources/14-resources-for-finance-}
\url{knowledge-extraction}, October 2022.} on stock markets, tickers and currencies. In addition, we search for words such as \textit{company}, \textit{enterprise}, \textit{manufacturer} and \textit{shareholder}, which may refer to an asset. We then detect dependencies between these key terms. Next, we replace all references to stock markets, assets, asset abbreviations and currencies with the tags \textsc{stock}, \textsc{ticker}, \textsc{ticker\_abr} and \textsc{currency}, respectively. Using the same lexica, we also replace financial terms and abbreviations with the tag \textsc{fin\_abr}.

A search is also made for capitalised proper names in the news items. Initially, these names are checked in the above and replaced by \textsc{fin\_abr}, \textsc{ticker\_abr} and \textsc{ticker} tags, as appropriate, when there is a match. In the absence of a match, they are replaced by category tags taken from an entity recognition database.

We homogenise numerical values and dates, and group dates and times under the tag \textsc{date} and quantitative terms under the tag \textsc{num} using a Name Entity Recogniser (\textsc{ner}, see Section \ref{reldataset}). Table \ref{tab:beforeafterpreproOriginal} provides a complete example of co-reference resolution and tag processing with detection of financial terms, homogenisation and replacement using the tools specified in Section \ref{reldataset}.

\begin{table*}[!htbp]
\centering
\caption{\label{tab:beforeafterpreproOriginal} Example of news content before and after applying financial term detection, homogenisation and replacement procedures.}
\begin{tabular}{cc}
\toprule
\textbf{ Before} & \textbf{After} \\\hline
\begin{tabular}[c]{@{}p{5.5cm}@{}}
Verizon Communications (\textsc{nyse:vz}) is proving to be a stable but undervalued company in the market today. In fact, \textsc{vz} stock is worth at least 55\% more than \textbf{\textsc{vz} stock} price today using an analysis of \textbf{\textsc{vz} stock} dividend yield, \textbf{\textsc{vz} stock} own P/E ratio history, and a comparison with \textbf{\textsc{vz} stock} peers.
\end{tabular} & 
\begin{tabular}[c]{@{}p{5.5cm}@{}}
\textbf{\textsc{ticker}} (\textbf{\textsc{stock:ticker\_abr}}) is proving to be a stable but undervalued company in the market today. In fact, \textbf{\textsc{ticker\_abr}} stock is worth at least \textbf{\textsc{num}} more than \textbf{\textsc{ticker\_abr} stock} price today using an analysis of \textbf{\textsc{ticker\_abr} stock} dividend yield, \textbf{\textsc{ticker\_abr} stock} own \textbf{\textsc{fin\_abr}} ratio history, and a comparison with \textbf{\textsc{ticker\_abr} stock} peers.
\end{tabular}
\\\bottomrule
\end{tabular}
\end{table*}

\subsection{Relevant text detection with \textsc{lda} topic modelling}
\label{features}

\textsc{lda} \cite{Jelodar2019} is an unsupervised algorithm that identifies different topics in a particular document, but it can be generalised to unknown documents if they belong to the same domain and share similar context and structure \cite{Gupta2021}. We use it to differentiate between relevant and less relevant information in segments (``documents'' in this section) produced from financial news content. Our goal is thus to discover relevant information in financial news by separating it from non-relevant information. Note that this type of news has a rather characteristic structure where relevant information is often presented along with precise contextual data and expression patterns, unlike other conventional news items.

Therefore, we employ a Dirichlet distribution with two topics (Equation \ref{eq:lda}). The training algorithm iterates to minimise the number of topics per word and document.

\begin{equation}
\label{eq:lda}
\begin{split}
P(Z,W,\theta,\phi;\alpha,\beta) =
\prod_{j=1}^{M} P(\theta_j;\alpha)
\times
\prod_{i=1}^{2} P(\phi_i;\beta)
\times \prod_{t=1}^{N} P(Z_{j,t} \rvert \theta_j)
P(W_{j,t} \rvert \phi_{Z_{j,t}})
\end{split}
\end{equation}

\begin{itemize}
\item \texttt{Z} is the set of target topics, two in this work.
\item \texttt{W} is the set of words (once stop-words\footnote{Available at \url{bit.ly/3yzvXqJ}, October 2022.} are discarded), with size $N$.
\item \texttt{M} is the size of the document collection.
\item $\alpha$ and $\beta$ are the symmetric smoothing hyper-parameters to avoid discarding one of the topics due to zero intra-document or intra-corpus topic occurrences. Both hyper-parameters are initialised randomly. Specifically, $\alpha$ and $\beta$ are the topic-document and word-topic densities, respectively. Lower values of $\alpha$ and $\beta$ reduce the variability of topic assignment to specific documents and words.
\item $P(\theta_j;\alpha)$ and $P(\phi_i;\beta)$ are the topic-documents and word-topics Dirichlet distributions, respectively.
\item $P(Z_{j,t} \rvert \theta_j)$ and $P(W_{j,t} \rvert \phi_{Z_{j,t}})$ are the topic-documents and word-topics multinomial distributions, respectively.
\end{itemize}

During \textsc{lda} model training, by modifying $\alpha$ and $\beta$, ($i$) topics are assigned randomly to the words in each document, then, ($ii$) the algorithm iterates across the word-topic pairs in different documents generating new assignments and accepting them if they decrease the number of topics per word and document.

The algorithm converges when it finds a solution that minimises the number of intra-document topics and topics per word. Alternatively, an iteration limit can be set. Ultimately, the resulting assignment of words to topics can be used to define a criterion for detecting topics in new text. To this end, when a new sentence is presented to the algorithm, the step ($ii$) is repeated by also taking into account the words in the new sentence. The score of a sentence for a given topic is the number of words of that topic in the sentence divided by the length of the sentence in words. In principle, the sentence is considered to belong to the topic with the largest score. Note that in this estimation, the algorithm is started using the distribution with the best hyper-parameters $\alpha$ and $\beta$ when the training algorithm terminates.

The capability of the system to differentiate between relevant and non-relevant information in the resulting topics is related to two combined effects of data conditioning in previous stages. First, TextTiling groups text by the different expression patterns that the authors of financial news tend to use in relevant and non-relevant text. Second, co-reference resolution and tag processing create a higher density of certain tags in relevant text. For this reason, as a practical contribution, we defined a topic score $\rho$ that represents the density of significant tags \textsc{stock}, \textsc{ticker}, \textsc{currency} and \textsc{fin\_abr} in financial news content, which is computed as the percentage of significant tags in a topic divided by the total number of tags in the whole data set. The topic with the highest $\rho$ value is considered relevant. Furthermore, to improve the precision of the \textsc{lda} algorithm in detecting relevant text (that is, its ability to avoid false positives) we introduced another practical contribution: an \textsc{lda} score threshold for accepting a sentence as relevant. It is computed as the minimum value of the configurable parameter $\delta$ and the mean value of the topic scores of the relevant sentences in the same segment.

Table \ref{tab:lda_results_example} shows an example of relevant information detection, together with some sentences on the relevant topic from the same segment and the corresponding \textsc{lda} classification scores. The mean value of the scores is 0.878\footnote{Note that this mean value is computed with all relevant sentences in the segment, only some of them are contained in Table \ref{tab:lda_results_example}.}. Thus, assuming that $\delta=0.8$, even though the third sentence belongs to the relevant topic, the system would consider it irrelevant.

\begin{table*}[!htbp]
\small
\centering
\caption{Example of \textsc{lda} analysis of relevant information.}
\begin{tabular}{cc}
\toprule \textbf{Text} & \textbf{\textsc{lda} score} \\ \hline
\begin{tabular}[c]{@{}p{10cm}@{}}{\textbf{\textsc{ticker}} (\textbf{\textsc{stock:ticker\_abr}}) is proving to be a stable but undervalued company in the market today. }\end{tabular} & 0.847\\\\

\begin{tabular}[c]{@{}p{10cm}@{}}{In fact, \textbf{\textsc{ticker\_abr}} stock is worth at least \textbf{\textsc{num}} more than \textbf{\textsc{ticker\_abr} stock} price today using an analysis of \textbf{\textsc{ticker\_abr} stock} dividend yield, \textbf{\textsc{ticker\_abr} stock} own \textbf{\textsc{fin\_abr}} ratio history, and a comparison with \textbf{\textsc{ticker\_abr} stock} peers.}\end{tabular} & 0.948\\\\

\begin{tabular}[c]{@{}p{10cm}@{}}{\textbf{\textsc{ticker}} reported “boring” earnings, according to Barron’s magazine for \textbf{\textsc{fin\_abr}} on \textbf{\textsc{date}}.}\end{tabular} & \bf{0.571<0.8}\\\\

\begin{tabular}[c]{@{}p{10cm}@{}}{For example, \textbf{\textsc{ticker}} said that \textbf{\textsc{ticker}} earnings on a non-\textbf{\textsc{fin\_abr}} adjusted basis was \textbf{\textsc{num}} per share.}\end{tabular} & 0.825\\\\

\begin{tabular}[c]{@{}p{10cm}@{}}{However, \textbf{\textsc{ticker}} generated higher cash flow.}\end{tabular} & 0.870\\\\

\begin{tabular}[c]{@{}p{10cm}@{}}{Moreover, this free cash flow (\textbf{\textsc{fin\_abr}}) in the first half was \textbf{\textsc{num}} billion, an increase of \textbf{\textsc{num}} percent year over year.}\end{tabular} & 0.897\\\\

\begin{tabular}[c]{@{}p{10cm}@{}}{\textbf{\textsc{ticker}} trades for a paltry \textbf{\textsc{num}} times this year’s expected earnings and just \textbf{\textsc{num}} times next year.}\end{tabular} & 0.934\\
\bottomrule
\end{tabular}
\label{tab:lda_results_example}
\end{table*}

\subsection{Temporal analysis}
\label{temporal_analysis}

Table \ref{tab:features} shows the set of temporal features used to train the \textsc{ml} temporal analysis model. The focal point of this analysis is the use of verbs when referring to stock markets, assets and currencies, but, unlike previous works, we consider them to be part of the semantic context.

\begin{table*}[!htbp]
\centering
\caption{\label{tab:features} Temporal features in the \textsc{ml} temporal analysis model.}
\begin{tabular}{ll}
\toprule
\textbf{Name} & \textbf{Description} \\\hline

\begin{tabular}[c]{@{}p{4cm}@{}} \{Pst,Prs,Fut\}DepSub \end{tabular} & \begin{tabular}[c]{@{}p{7.5cm}@{}} Number of \{past,present,future\} tense verbs from the dependency analysis when the asset is the subject of the clause.\end{tabular}\\

\begin{tabular}[c]{@{}p{4cm}@{}} GlobalDepSub \end{tabular} & \begin{tabular}[c]{@{}p{7.5cm}@{}} Global temporality by majority voting from the dependency analysis when the asset is the subject of the clause.\end{tabular}\\

\begin{tabular}[c]{@{}p{4cm}@{}} \{Pst,Prs,Fut\}DepSubObj \end{tabular} & \begin{tabular}[c]{@{}p{7.5cm}@{}} Number of \{past,present,future\} tense verbs from the dependency analysis when the asset is the subject or object of the clause.\end{tabular}\\

\begin{tabular}[c]{@{}p{4cm}@{}} GlobalDepSubObj \end{tabular} & \begin{tabular}[c]{@{}p{7.5cm}@{}} Global temporality by majority voting from the dependency analysis when the asset is the subject or object of the clause.\end{tabular}\\

\begin{tabular}[c]{@{}p{4cm}@{}} \{Pst,Prs,Fut\}ProxSub \end{tabular} & \begin{tabular}[c]{@{}p{7.5cm}@{}} Number of \{past,present,future\} tense verbs from the proximity analysis when the asset is the subject of the clause.\end{tabular}\\

\begin{tabular}[c]{@{}p{4cm}@{}} GlobalProxSub \end{tabular} & \begin{tabular}[c]{@{}p{7.5cm}@{}} Global temporality by majority voting from the proximity analysis when the asset is the subject of the clause.\end{tabular}\\

\begin{tabular}[c]{@{}p{4cm}@{}} \{Pst,Prs,Fut\}ProxSubObj \end{tabular} & \begin{tabular}[c]{@{}p{7.5cm}@{}} Number of \{past,present,future\} tense verbs from the proximity analysis when the asset is the subject or object of the clause.\end{tabular}\\

\begin{tabular}[c]{@{}p{4cm}@{}} GlobalProxSubObj \end{tabular} & \begin{tabular}[c]{@{}p{7.5cm}@{}} Global temporality by majority voting from the proximity analysis when the asset is the subject or object of the clause.\end{tabular}\\
\bottomrule
\end{tabular}
\end{table*}

Within each relevant sentence, we perform a dependency analysis to link verbs to stock markets, assets and currencies and a proximity analysis based on the proximity between the verbs and the key elements identified. The system measures the distance of a term to the nearest verb as the number of intermediate words in both directions. For both analyses (dependency and proximity), we consider whether assets are the subjects or objects of their clauses. For each feature, we estimate the verb tense by majority voting among past, present and future tenses (in case of a tie, the future tense prevails).

Algorithms \ref{alg:dependency} and \ref{alg:proximity} describe the generation of the temporal features of a segment based on the dependency (Algorithm \ref{alg:dependency}) and the proximity (Algorithm \ref{alg:proximity}) analysis of the corresponding sentences. Both algorithms have linear time complexity O($n\cdot d$) owing to the \textit{d} independent loops with a limited number of \textit{n} elements (subject and object sentences in our case). In the particular scenario in which the number of subject sentences equals the number of object sentences, for each loop and each tense, time complexity is O($2\cdot3\cdot n$) for 2 analyses (subject and object) and 3 tenses (past, present, future). Table \ref{news_temporality} shows an example of the outcome. Note the financial terms and associated verbs. For this segment, for example, features \texttt{FutDepSubObj} and \texttt{FutProxSubObj} in Table \ref{tab:features} are set to 1. 

\begin{algorithm*}[!htbp]
 \caption{\label{alg:dependency}: {\bf Dependency analysis}}
 \begin{algorithmic}[0]
 \Function{dependency\_analysis}{text}
 \State \{Pst,Prs,Fut\}DepSub = \{Pst:0,Prs:0,Fut:0\}
 \State \{Pst,Prs,Fut\}DepSubObj = \{Pst:0,Prs:0,Fut:0\}
 \State DepSub\_text = parsing\_subject(text)
 \State DepSubObj\_text = parsing\_subject\_object(text)
 \For{tense in [Pst,Prs,Fut]}
 \State \{Pst,Prs,Fut\}DepSub.add(tense,
 \State get\_number\_verb\_tense(DepSub\_text,tense))
 \EndFor
 \State GlobalDepSub = majority(\{Pst,Prs,Fut\}DepSub)
 \For{tense in [Pst,Prs,Fut]}
 \State \{Pst,Prs,Fut\}DepSubObj.add(tense,
 \State get\_number\_verb\_tense(DepSubObj\_text,tense))
 \EndFor
 \State GlobalDepSubObj = majority(\{Pst,Prs,Fut\}DepSubObj)
 \EndFunction
 \end{algorithmic}
\end{algorithm*} 

\begin{algorithm*}[!htbp]
 \caption{\label{alg:proximity}: {\bf Proximity analysis}}
 \begin{algorithmic}[0]
 \Function{proximity\_analysis}{text}
 \State \{Pst,Prs,Fut\}ProxSub = \{Pst:0,Prs:0,Fut:0\}
 \State \{Pst,Prs,Fut\}ProxSubObj = \{Pst:0,Prs:0,Fut:0\}
 \State ProxSub\_text = parsing\_subject(text)
 \State ProxSubObj\_text = parsing\_subject\_object(text)
 \For{tense in [Pst,Prs,Fut]}
 \State \{Pst,Prs,Fut\}ProxSub.addNearest(tense,
 \State get\_number\_verb\_tense(ProxSub\_text,tense))
 \EndFor
 \State GlobalProxSub = majority(\{Pst,Prs,Fut\}ProxSub)
 \For{tense in [Pst,Prs,Fut]}
 \State \{Pst,Prs,Fut\}ProxSubObj.addNearest(tense,
 \State get\_number\_verb\_tense(ProxSubObj\_text,tense))
 \EndFor
 \State GlobalProxSubObj = majority(\{Pst,Prs,Fut\}ProxSubObj)
 \EndFunction
 \end{algorithmic}
\end{algorithm*}

In addition to the temporal features, we also consider textual and numerical features in the \textsc{ml} temporal analysis model. The textual features are char-grams, word-grams and word tokens ($n$-grams within word boundaries), whose parameter ranges are selected by combinatorial searching. The two numerical features are the number of numerical values (excluding percentages) and the number of percentages in the news content. 

After empirical tests with diverse \textsc{ml} algorithms, we chose a Linear Support Vector Classifier (\textsc{svc}) to estimate the temporality (past, present, future) of a segment (see our previous work \cite{Garcia-Mendez2022}). Before training the \textsc{svc}, we pre-processed the clauses in the financial news by converting text to lower case, and removing punctuation marks and non-Unicode characters such as accents and symbols.

Finally, Algorithm \ref{alg:flow} shows the logical flow of the proposed solution.

\begin{algorithm*}[!htbp]
 \caption{\label{alg:flow}: {\bf Solution pipeline}}
 \begin{algorithmic}[0]

 \Function{relevant\_forecast\_detection}{financial\_news}
 
 \State segments = TextTiling(financial\_news) \textbf{\small \%Multi-paragraph topic segmentation}

 \For{s in segments}

  \State NeuralCoref(s) \textbf{\small \%Co-reference resolution}
  
  \State \textbf{\small \%Financial terms detection, homogenisation and replacement}
  \State replaceAll(s,stock\_dict,"STOCK")
  
  \State replaceAll(s,asset\_dict,"TICKER")
  
  \State keywords = ["company","enterprise","manufacturer","shareholder"]
  \State replaceAll(s,keywords,"TICKER")
  
  \State replaceAll(s,assetsAbr\_dict,"TICKER\_ABR")
  
  \State replaceAll(s,currency\_dict,"CURRENCY")
  
  \State replaceAll(s,financial\_dict,"FIN\_ABR")
 
  \State NER(s)

 \EndFor

  \State topic\_modelling = LDA(segments) \textbf{\small \%Relevant text detection with LDA topic modelling}
 
 \State relevant\_topic=0
 
 \State highest\_$\rho$ = 0
 
 \For{topic in topics}
 
  \State topic\_$\rho$ = max(topic\_modelling.filterTopic[topic].$\rho$)
  
  \If{topic\_$\rho$ > highsest\_$\rho$}
   \State relevant\_topic = topic
   \State highsest\_$\rho$ = topic\_$\rho$
   
  \EndIf

 \EndFor 
 
 \State relevant = topic\_modelling.filterTopic[relevant\_topic]
  
 \For{sentence in relevant}
 
  \If{sentence.score < $\delta$}
   
   \State relevant.remove(sentence)
   
  \EndIf

 \EndFor
 
 \State \textbf{\small \%Temporal analysis}
 \For{s in segments}
 
  \State to\_lowercase(s)
  
  \State remove\_punctuation\_non-Unicode(s)
  
  \State dependency = dependency\_analysis(s)
 
  \State proximity = proximity\_analysis(s)
 
  \State textual = compute\_chargrams\_wordgrams\_tokens(s)
  
  \State numerical = count\_numerical\_values\_percentages(s)
  
 \EndFor

 \State trainSVC(dependency,proximity,textual,numerical)
 
 \EndFunction
 \end{algorithmic}
\end{algorithm*}

\section{Experimental results}
\label{sec:results}

In this section, we describe the experimental data set and the performance of our system for the detection of relevant financial events. Well-known state-of-the-art metrics were used for the evaluation: Alpha-reliability and accuracy \cite{Giannantonio2010}, and the Recall-Oriented Understudy for Gisting Evaluation (\textsc{rouge}) metric\footnote{Available at \url{https://github.com/pltrdy/rouge}, October 2022.} \cite{Mohamed2019,Sanchez-Gomez2018,El-Kassas2020}. Comparisons of the proposed solution with a rule-based baseline and a supervised extraction approach are also provided.
 
\subsection{Experimental setting}
\label{reldataset}

The experiments were performed on a computer with the following specifications:

\begin{itemize}
 \item Operating system: Ubuntu 18.04.2 \textsc{lts} 64 bits
 \item Processor: Intel\@Core i9-9900K 3.60 GHz 
 \item RAM: 32 GB DDR4 
 \item Disk: 500 GB (7200 rpm SATA) + 256 GB SSD
\end{itemize}

\begin{table*}[!htbp]
\centering
\caption{\label{news_temporality} Example of news entry after applying dependency and proximity analyses.}
\begin{tabular}{cc}
\toprule
\textbf{Dependency analysis} & \textbf{Proximity analysis} \\\hline
\begin{tabular}[c]{@{}p{5.5cm}@{}}
\textbf{\textsc{ticker (stock:ticker\_abr)} is proving to be} a stable but undervalued company in the market today. In fact, \textbf{\textsc{ticker\_abr}} stock \textbf{is} worth at least 55\% more than \textsc{ticker\_abr} stock price today using an analysis of \textsc{ticker\_abr} stock dividend yield, \textsc{ticker\_abr} stock own \textsc{fin\_abr} ratio history, and a comparison with \textsc{ticker\_abr} stock peers.
\textsc{ticker} reported “boring” earnings, according to Barron’s magazine for Q2 on July 24. But Barron’s magazine for Q2 says “boring is good” in this market. They point out that the communications business is not a bad place to be in a pandemic.
For example, \textbf{\textsc{ticker} said} that \textsc{ticker} earnings on a non-\textsc{fin\_abr} adjusted basis was \$1.18 per share. This was only 3\% lower than a year ago. On an adjusted \textsc{fin\_abr} (earnings before interest, taxes, depreciation, and amortization) basis, taxes, depreciation, and amortization) cash flow was down 5\%.
However, \textbf{\textsc{ticker} generated} higher cash flow. For example, first-half 2020 cash flow from operations of \$23.6 billion, an increase of \$7.7 billion from first-half of 2019. This represents a huge increase of over 206\%. Moreover, this free cash flow (\textsc{ticker\_abr}) in the first half was \$13.7 billion, an increase of 74.1 percent year over year.
But consider this. \textsc{ticker} trades for a paltry 12.3 times this year’s expected earnings and just 12 times next year.
\end{tabular} & 
\begin{tabular}[c]{@{}p{5.5cm}@{}}
\textbf{\textsc{ticker (stock:ticker\_abr)} is proving to be} a stable but undervalued company in the market today. In fact, \textbf{\textsc{ticker\_abr}} stock \textbf{is} worth at least 55\% more than \textsc{ticker\_abr} stock price today using an analysis of \textsc{ticker\_abr} stock dividend yield, \textsc{ticker\_abr} stock own \textsc{fin\_abr} ratio history, and a comparison with \textsc{ticker\_abr} stock peers.
\textbf{\textsc{ticker} reported} “boring” earnings, according to Barron’s magazine for Q2 on July 24. But Barron’s magazine for Q2 says “boring is good” in this market. They point out that the communications business is not a bad place to be in a pandemic.
For example, \textbf{\textsc{ticker} said} that \textsc{ticker} earnings on a non-\textsc{fin\_abr} adjusted basis was \$1.18 per share. This was only 3\% lower than a year ago. On an adjusted \textsc{fin\_abr} (earnings before interest, taxes, depreciation, and amortization) basis, taxes, depreciation, and amortization) cash flow was down 5\%.
However, \textbf{\textsc{ticker} generated} higher cash flow. For example, first-half 2020 cash flow from operations of \$23.6 billion, an increase of \$7.7 billion from first-half of 2019. This represents a huge increase of over 206\%. Moreover, this free cash flow (\textbf{\textsc{ticker\_abr}}) in the first half \textbf{was} \$13.7 billion, an increase of 74.1 percent year over year.
But consider this. \textsc{ticker} trades for a paltry 12.3 times this year’s expected earnings and just 12 times next year.
\end{tabular}
\\\bottomrule
\end{tabular}
\end{table*}

Regarding the implementation, we took the following decisions: 

\begin{itemize}

\item Segmentation (Section \ref{textiling}): as previously mentioned, we applied the TextTiling algorithm\footnote{Available at \url{https://www.nltk.org/\_modules/nltk/tokenize/texttiling.html}, October 2022.}. For this first stage, we used the default parameters of the classifier (see Listing \ref{textiling_parameters}). Stop-words were also removed.

\item Co-reference detection (Section \ref{section:coref}): we used the \textsc{nn} implementation of the NeuralCoref library\footnote{Available at \url{https://github.com/huggingface/neuralcoref}, October 2022.}, which is a pipelined extension of the spaCy library\footnote{Available at \url{https://spacy.io}, October 2022.} to solve co-reference groups in blocks of text. Specifically, we used the pre-trained word embedding statistical model for English with its default features. Surprisingly, in our analysis, we noticed that NeuralCoref did not always detect the indispensable word \textit{stock}. To circumvent this problem, following a trial and error process, we replaced \textit{stock} with \textit{index} before co-reference resolution.

\begin{lstlisting}[frame=single,caption={Configuration parameter ranges of the TextTiling algorithm.}, label={textiling_parameters}]
W: 20
K: 10
similarity_method: block comparison
stopwords_list: nltk
smoothing_width: 2
smoothing_rounds: 1
\end{lstlisting}

\item Tag processing (Section \ref{processing}): 
 we employ the Freeling library\footnote{Available at \url{http://nlp.lsi.upc.edu/freeling/node/1}, October 2022.} to detect dependencies between terms. Capitalised proper names are detected with the spaCy\footnote{Available at \url{https://spacy.io}, October 2022.} library. If the names are not in our lexica, their categories are selected from spaCy EntityRecognizer tool\footnote{Available at \url{https://spacy.io/api/top-level\#spacy.explain} and \url{https://github.com/explosion/spaCy/blob/master/
 spacy/glossary.py}, October 2022.}.
 Depending on the category, the names are replaced by the following tags: \textsc{money}, \textsc{person}, \textsc{norp} (nationalities, religions or political groups), \textsc{org} (organisations, companies), \textsc{product}, \textsc{event}, and \textsc{work of art} (titles of artworks). Moreover, all elements recognised as \textsc{loc}, \textsc{fac} (buildings) and \textsc{gpe} (countries, cities, states) are grouped under the tag \textsc{loc} (locations). 
 Numerical values and dates are detected using EntityRecognizer. Terms recognised as \textsc{date} and \textsc{time} are grouped under the tag \textsc{date}, while those recognised as \textsc{percent}, \textsc{cardinal} and \textsc{quantity} are grouped under the tag \textsc{num}.

\item Relevant text detection with \textsc{lda} topic modelling (Section \ref{features}): we employed the LdaMulticore module from the gensim Python library\footnote{Available at \url{https://pypi.org/project/gensim}, October 2022.}. Listing \ref{ldamulticore_parameters} shows the configuration parameters used to train the model. Through repeated trials, we tuned the algorithm to 50 training \texttt{passes}, \texttt{alpha} to \texttt{symmetric} and \texttt{beta} to \texttt{asymmetric}. Finally, we set $\delta=0.8$.

\begin{itemize}
 \item \texttt{Numtopics} is the number of latent topics to be extracted.
 
 \item \texttt{Passes} is the number of passes to be applied during training.
 
 \item \texttt{Random state} is a useful seed for reproducibility.
 
 \item \texttt{Alpha} represents an a-priori belief about document-topic distribution, that is, prior to selection strategies. Its feasible values are: (\textit{i}) \texttt{scalar} for symmetric document-topic distribution, (\textit{ii}) \texttt{symmetric} to use a fixed symmetric distribution of $1.0/num topics$, and (\textit{iii}) \texttt{asymmetric} to use a fixed normalised asymmetric distribution of $1.0/(topic index + sqrt(num topics))$.

 \item \texttt{Beta} is an a-priori belief on topic-word distribution. It has the same feasible values as \texttt{alpha}.
 
\end{itemize}

\item Temporal analysis (Section \ref{temporal_analysis}): Freeling is used to tag assets as subjects or objects of their clauses to obtain the corresponding temporal features. We used the \textsc{svc} implementation from the Scikit-Learn Python library\footnote{Available at \url{https://scikit-learn.org/stable/supervised
\_learning.html\#supervised-learning}, October 2022.}. Regarding the parameterisation of the char-grams, word-grams and word tokens textual features, we applied a GridSearchCV\footnote{Available at \url{https://scikit-learn.org/stable/modules/
generated/sklearn.model\_selection.GridSearchCV.html}, October 2022.} combinatorial search from Scikit-Learn within the ranges in our prior related work \cite{Garcia-Mendez2022}.
The final choices were \texttt{maxdf} = 0.30, \texttt{mindf} = 0, \texttt{ngramrange} = (2,4) and \texttt{maxfeatures} = 10000. 

\begin{itemize}
 \item \texttt{Mindf} and \texttt{maxdf} are used to ignore terms with a lower (cut-off) and higher (corpus-specific stop words) document frequency than the given threshold, respectively.
 
 \item \texttt{Ngramrange} indicates the lower and upper boundary for the extraction of word $n$-grams.
 
 \item \texttt{Maxfeatures} represents the number of features considered for the best split.
 
 \end{itemize}

To select the best features for the temporal analysis across the whole set (temporal, textual and numerical features), we used SelectPercentile\footnote{\label{note1}Available at \url{https://scikit-learn.org/stable/modules/
feature\_selection.html}, October 2022.} from Scikit-Learn with the $\chi^2$ score function and 80th percentile threshold. The hyper-parameters of the \textsc{svc} were tuned using GridSearchCV with 10-fold cross validation within the ranges in Listing \ref{configuration_parameters_svn}. The optimal hyper-parameter values used were \texttt{C} = 0.001, \texttt{classweight} = balanced, \texttt{loss} = \texttt{squared\_hinge}, \texttt{maxiter} = 1500, \texttt{multiclass} = ovr, \texttt{penalty} = l2, \texttt{tol} = $10^{-9}$. Finally, the \textsc{svc} was evaluated by 10-fold cross validation using 600 sentences from financial news. This auxiliary data set is similar in size to other sets used in the literature \cite{Li2018,Long2019,Al-Smadi2019,Zhang2020,DeOliveiraCarosia2021,Xie2021} and did not belong to the experimental data set described in Section \ref{reldataset} that was used to detect relevant information, and was independently annotated. The \textsc{svc} attained 80.21\% precision and 80.40\% recall for the auxiliary set.

 \begin{itemize}

 \item \texttt{C} is the regularisation parameter.
 
 \item \texttt{Classweight} is used to set the parameter \texttt{C} of the classes. If not given, all classes are assumed to have weight one. The \texttt{balanced} mode automatically adjusts weights in a manner that is inversely proportional to class frequencies in the input data.
 
 \item \texttt{Loss} represents the loss function.
 
 \item \texttt{Maxiter} is the hard limit on iterations, or -1 for no limit.
 
 \item \texttt{Multiclass} represents the one-vs-one scheme.
 
 \item \texttt{Penalty} represents the penalty for the model.
 
 \item \texttt{Tol} represents the tolerance for the stopping criterion.
 
\end{itemize}

\end{itemize}

\begin{lstlisting}[frame=single,caption={Configuration parameter ranges for the \textsc{lda} model.}, label={ldamulticore_parameters}]
numtopics: 2
passes: (25,50,75,100)
randomstate: 1
alpha|beta: (0.01,0.31,0.61,0.91,symmetric,asymmetric)
\end{lstlisting}

\begin{lstlisting}[frame=single,caption={Configuration parameter ranges for \textsc{svc}.}, label={configuration_parameters_svn}]
C: (0.0001, 0.0005, 0.001, 0.01)
classweight: [None,balanced]
loss: squared_hinge
maxiter: [500, 1000, 1500]
multiclass: ovr
penalty: l2
tol: (0.000000001, 0.00000001, 0.0000001, 0.000001)
\end{lstlisting}

\subsection{Experimental data set}
\label{dataset}

Our experimental data set was composed of 2,158 news pieces (average length of 27.98 sentences and 537.24 words). As previously mentioned, the pieces were automatically extracted with a script from popular and prestigious financial websites between 1st October 2018 and 1st October 2020. We filtered the news pieces to keep those that mentioned at least one of the stocks in our financial lexica\footref{gtiresources}. For text processing purposes, double spaces, line breaks and tabs were replaced by a single space. Finally, we removed, \textsc{url}s, images and graphics and kept the date and author information.

Each entry in the resulting data set is composed of an identifier, a title, content, author information, source and date of publication. The entries are comparable in size to those described in previous \textsc{ke} \cite{Li2018,Long2019,Al-Smadi2019,Zhang2020,DeOliveiraCarosia2021,Xie2021} and \textsc{lda} works \cite{Park2019}.

A brief descriptive analysis of the data set is given in Table \ref{tab:dataset}.

\begin{table}[htb]
\centering
\caption{\label{tab:dataset} Descriptive analysis of the experimental data set.}
\begin{tabular}{clllll}
\cmidrule{2-6}
& \multicolumn{1}{c}{\bf Source} & \multicolumn{1}{c}{\bf Authors} & \multicolumn{1}{c}{\bf News} & \multicolumn{1}{c}{\bf Avg. chars} & \multicolumn{1}{c}{\bf Avg. words}\\
\cmidrule{2-6}
& Benzinga\tablefootnote{Available at \url{https://www.benzinga.com}, October 2022.} & 152 & 400 & 3325.21 & 536.295\\
& The Motley Fool\tablefootnote{Available at \url{https://www.fool.com}, October 2022.} & 75 & 298 & 3537.58 & 576.69\\
& Markets Insider\tablefootnote{Available at \url{https://markets.businessinsider.com}, October 2022.} & 13 & 223 & 5270.50 & 758.61\\
& TipRanks\tablefootnote{Available at \url{https://www.tipranks.com}, October 2022.} & 4 & 210 & 1505.61 & 235.84\\
& Business Standard\tablefootnote{Available at \url{https://www.business-standard.com}, October 2022.} & 22 & 198 & 2615.02 & 430.00\\
& Investorplace\tablefootnote{Available at \url{https://investorplace.com}, October 2022.} & 42 & 158 & 4967.15 & 827.32\\
& Iwatch Markets\tablefootnote{Available at \url{https://iwatchmarkets.com}, October 2022.} & 2 & 154 & 555.51 & 95.24\\
& Investing Daily\tablefootnote{Available at \url{https://www.investingdaily.com}, October 2022.} & 2 & 150 & 6319.58 & 1011.69\\
& Investopedia\tablefootnote{Available at \url{https://www.investopedia.com}, October 2022.} & 17 & 140 & 4462.16 & 703.70\\
& Gurufocus\tablefootnote{Available at \url{https://www.gurufocus.com}, October 2022.} & 20 & 49 & 4815.94 & 783.65\\
& CNN Business\tablefootnote{Available at \url{https://edition.cnn.com/markets}, October 2022.} & 32 & 49 & 4507.49 & 745.49\\
& Market Watch\tablefootnote{Available at \url{https://www.marketwatch.com}, October 2022.} & 28	& 47 & 4352.68 & 721.57\\
& Seeking Alpha\tablefootnote{Available at \url{https://seekingalpha.com}, October 2022.} & 13 & 33 & 783.67 & 124.45\\
& CNBC\tablefootnote{Available at \url{https://www.cnbc.com}, October 2022.} & 4 & 31 & 1962.16 & 338.87\\
& IOL\tablefootnote{Available at \url{https://www.iol.co.za}, October 2022.} & 6 & 18 & 1046.00 & 169.22\\\midrule
\bf Total & 15 & 431\tablefootnote{Without duplicates.} & 2158 & 3335.08 & 537.24\\
\bottomrule
\end{tabular}
\end{table}

The texts were annotated by five \textsc{nlp} scientists from the atlanTTic Research Centre for Telecommunication Technologies at the University of Vigo. Manual annotations included relevant texts, asset identifiers and prediction/forecast texts. A number of guidelines were agreed on to enhance consistency (\textit{e.g.}, bold font for relevant text, italics for asset identifiers and underlining for prediction/forecast text). Table \ref{news_use_case_annotated} shows an example of an annotated news item from the experimental data set. We used the annotated asset identifiers to improve the content of our financial lexica\footref{gtiresources} which increased by 3.95\%. 

\begin{table*}[!htbp]
\centering
\caption{\label{news_use_case_annotated} Example of annotated news entry.}
\begin{tabular}{c}
\toprule
\begin{tabular}[c]{@{}p{11cm}@{}}
Verizon Communications (\textsc{nyse:vz}) is proving to be a stable but undervalued company in the market today. \textbf{In fact, \textit{VZ} stock is worth at least 55\% more than its price today using an analysis of its dividend yield, its own P/E ratio history, and a comparison with its peers.}
The company reported “boring” earnings, according to Barron’s magazine for Q2 on July 24. But the magazine says “boring is good” in this market. They point out that the communications business is not a bad place to be in a pandemic. For example, Verizon said that its earnings on a non-GAAP adjusted basis was \$1.18 per share. This was only 3\% lower than a year ago. On an adjusted \textsc{ebitda} (earnings before interest, taxes, depreciation, and amortization) basis, its cash flow was down 5\%. However, \textit{Verizon} generated higher cash flow. For example, first-half 2020 cash flow from operations of \$23.6 billion, an increase of \$7.7 billion from first-half of 2019. This represents a huge increase of over 206\%. Moreover, its free cash flow (\textsc{fcf}) in the first half was \$13.7 billion, an increase of 74.1 percent year over year. But consider this. \textbf{\textit{Verizon} trades for a paltry 12.3 times this year’s expected earnings and \underline{just 12 times next year}[...]}
\end{tabular}
\\\bottomrule
\end{tabular}
\end{table*}

\subsection{Inter-agreement evaluation}
We evaluated inter-annotator agreement using two well-known state-of-the-art metrics: Alpha-reliability and accuracy.

Table \ref{coincidenceMatrix} shows the coincidence matrix of relevance across all annotators. The two components in the diagonal show the number of news sentences on which all the annotators agreed, while the other two components show the cases on which at least one annotator disagreed. Tables \ref{alphaMatrix} and \ref{accurayMatrix} show the Alpha-reliability and accuracy coefficients by pairs of annotators. The mean values were 0.552 and 0.861, respectively. Previous works have considered an Alpha-reliability value above 0.41 to be acceptable \cite{Rash2019,Seite2019,Salminen2019,Kilicoglu2021}. Inter-agreement accuracy was very high, at over 80\%.

\begin{table}[!htbp]
\centering
\caption{\label{coincidenceMatrix}Coincidence matrix for relevant text annotation.} 
\begin{tabular}{ccc}
\toprule
\multicolumn{1}{l}{} & \textbf{Relevant} & \textbf{Context}\\ \hline
\textbf{Relevant} & 2752.5 & 1561.5\\
\textbf{Context} & 1561.5 & 16584.5\\\bottomrule
\end{tabular}
\end{table}

\begin{table*}[!htbp]
\centering
\caption{\label{alphaMatrix}Inter-agreement Alpha-reliability of relevant text annotation by pairs of annotators (An.).}
\begin{tabular}{lccccc}\toprule
\multicolumn{1}{c}{} & \multicolumn{1}{c}{\bf An. 1} & \multicolumn{1}{c}{\bf An. 2} & \multicolumn{1}{c}{\bf An. 3} & \multicolumn{1}{c}{\bf An. 4} & \multicolumn{1}{c}{\bf An. 5} \\\hline
\multicolumn{1}{c}{\bf An. 1} & \multicolumn{1}{c}{-} & \multicolumn{1}{c}{0.569} & \multicolumn{1}{c}{0.550} & \multicolumn{1}{c}{0.605} & \multicolumn{1}{c}{0.629} \\
 \multicolumn{1}{c}{\bf An. 2} & 0.569 & \multicolumn{1}{c}{-} & \multicolumn{1}{c}{0.594} & \multicolumn{1}{c}{0.525} & \multicolumn{1}{c}{0.520} \\
 \multicolumn{1}{c}{\bf An. 3} & 0.550 & 0.594 & \multicolumn{1}{c}{-} & \multicolumn{1}{c}{0.436} & \multicolumn{1}{c}{0.506} \\
 \multicolumn{1}{c}{\bf An. 4} & 0.605 & 0.525 & 0.436 & \multicolumn{1}{c}{-} & \multicolumn{1}{c}{0.585} \\
 \multicolumn{1}{c}{\bf An. 5} & 0.629 & 0.520 & 0.506 & \multicolumn{1}{c}{0.585} & \multicolumn{1}{c}{-} \\\bottomrule
\end{tabular}
\end{table*}

\begin{table*}[!htbp]
\centering
\caption{\label{accurayMatrix}Inter-agreement accuracy of relevant text annotation by pairs of annotators (An.).}
\begin{tabular}{lccccc}\toprule
\multicolumn{1}{c}{} & {\bf An. 1} & {\bf An. 2} & {\bf An. 3} & {\bf An. 4} & {\bf An. 5} \\\hline
{\bf An. 1} & - & 0.862 & 0.844 & 0.891 & 0.895 \\
{\bf An. 2} & 0.862 & - & 0.853 & 0.859 & 0.855 \\
{\bf An. 3} & 0.844 & 0.853 & - & 0.818 & 0.838 \\
{\bf An. 4} & 0.891 & 0.859 & 0.818 & - & 0.894 \\
{\bf An. 5} & 0.895 & 0.855 & 0.838 & 0.894 & - \\\bottomrule
\end{tabular}
\end{table*}

\subsection{Discussion of the results}

Before applying our system for the automatic detection of relevant financial events, we first defined a simple rule-based system as a baseline. This system sets a relevance score by counting tickers, numbers and percentages and detecting future tenses using Freeling. As previously mentioned, relevant financial text has characteristic context data and expression patterns, and our goal was to determine whether a sophisticated technique such as \textsc{lda} would perform better than trivial rules.

The rule-based baseline approach is as follows. First, the text is segmented into sentences, and all references to stock markets, assets, asset abbreviations and currencies are replaced by the tags \textsc{stock}, \textsc{ticker}, \textsc{ticker\_abr} and \textsc{currency}, respectively. As indicated in Section \ref{processing}, financial terms and abbreviations are also replaced by the tag \textsc{fin\_abr}. Freeling is applied to detect percentages and numerical values, which are replaced by the tag \textsc{num}. Sentences containing a future tense as detected by Freeling are considered to refer to the future. In brief, a sentence is considered relevant if it contains at least one financial tag (\textsc{stock}, \textsc{ticker}, \textsc{ticker\_abr}, \textsc{currency} or \textsc{fin\_abr}) and at least one \textsc{num} tag, and predictive if the main verb is in future tense.

Drawing from related work on more powerful supervised extraction strategies \cite{Harb2008,Gottipati2018,Vermeer2019,Lopez-Ubeda2021}, we also applied a \textsc{svc} model\footnote{Note that a supervised approach was also used to estimate the temporality (past, present, future) of text in our proposal.} as a second comparison reference. The model was trained using manual annotations on relevant text, including predictions. Textual features were generated and hyper-parameter settings optimised as described in Section \ref{reldataset}.

Next, we evaluated our system by checking it against the annotated segments. To do this, we employed \textsc{rouge}, a widely used set of metrics for evaluating automatic text extraction performance based on overlapping $n$-grams. We used \textsc{rouge-l}, which measures the longest common sub-sequence between the system output and the annotated news. This \textsc{rouge} variant has been applied as a string matching algorithm to compute the similarity between two texts \cite{Gulden2019}. 

Tables \ref{rouge} and \ref{rougeTime} show the results obtained for the baseline systems and the proposed system. Even though the problem case is entirely novel (see Section \ref{contributions}), the results show that the application of sophisticated \textsc{nlp} techniques and \textsc{ke} algorithms such as those used in our solution results in improved extraction of relevance and temporality from financial news content.

Table \ref{rouge} shows the results for the detection of relevant information by the baseline systems and our system for the tags identified by the five annotators. Average values are also provided. In our tests, text was considered relevant when its score for a given topic doubled the score of the other topic\footnote{Note that there are only two topics under analysis in this work.}. The remaining text was considered to be less relevant or contextual information. The average \textsc{rouge-l} value across all annotators was 0.662, more than doubling the performance of the rule-based baseline approach. The manually intensive supervised extraction alternative was comparable to our unsupervised approach (in fact the former was often worse, depending on the annotator). This performance can be considered satisfactory in line with other works from the literature \cite{Sanchez-Gomez2018,El-Kassas2020,Hark2020,Alqaisi2020}. Table \ref{rougeTime} shows the results for the detection of relevant predictions/forecasts after the last \textsc{svc} classifier stage. The average \textsc{rouge-l} value in this case, 0.982, was excellent, with a significant improvement over the rule-based baseline reference of 0.713. 

Counterintuitively, the performance of the rule-based baseline in Table \ref{rouge} is worse because this approach gives relevance to text that contains quantitative data even if it corresponds to merely contextual information, such as past states of assets and stock markets. It underperformed our proposed solution by an average of 51\%. The level of agreement between our system and the annotators for the detection of predictions/forecasts (Table \ref{rougeTime}) was near perfect, with \textsc{rouge} values of more than 0.970 for all annotators. As expected, predictions and forecasts within relevant text are easier to detect than relevant text itself, explaining the lower \textsc{rouge} values in this second case, where the highest coefficient observed was 0.727 (for annotator 1).

\begin{table*}[!htbp]
\centering
\caption{\label{rouge}\textsc{rouge-l} measures for the detection of relevant text, by annotator (An.) and average.}
\begin{tabular}{lcccccc}\toprule
\multicolumn{1}{c}{} & \multicolumn{1}{c}{\bf An. 1} & \multicolumn{1}{c}{\bf An. 2} & \multicolumn{1}{c}{\bf An. 3} & \multicolumn{1}{c}{\bf An. 4} & \multicolumn{1}{c}{\bf An. 5} & \multicolumn{1}{c}{\bf Avg.} \\\hline

\multicolumn{1}{c}{¡\textbf{Rule-based baseline}} & \multicolumn{1}{c}{0.257} & \multicolumn{1}{c}{0.390} & \multicolumn{1}{c}{0.333} & \multicolumn{1}{c}{0.390} & \multicolumn{1}{c}{0.246} & \multicolumn{1}{c}{0.323} \\

\multicolumn{1}{c}{\textbf{Supervised system}} & \multicolumn{1}{c}{0.655} & \multicolumn{1}{c}{0.582} & \multicolumn{1}{c}{0.615} & \multicolumn{1}{c}{0.643} & \multicolumn{1}{c}{0.574} & \multicolumn{1}{c}{0.614} \\

\multicolumn{1}{c}{\textbf{Proposed system}} & \multicolumn{1}{c}{0.727} & \multicolumn{1}{c}{0.681} & \multicolumn{1}{c}{0.720} & \multicolumn{1}{c}{0.547} & \multicolumn{1}{c}{0.633} & \multicolumn{1}{c}{0.662} \\\bottomrule
\end{tabular}
\end{table*}

\begin{table*}[!htbp]
\centering
\caption{\label{rougeTime}\textsc{rouge-l} measures the detection of relevant text with predictions/forecasts, by annotator (An.) and average.}
\begin{tabular}{lcccccc}\toprule
\multicolumn{1}{c}{} & \multicolumn{1}{c}{\bf An. 1} & \multicolumn{1}{c}{\bf An. 2} & \multicolumn{1}{c}{\bf An. 3} & \multicolumn{1}{c}{\bf An. 4} & \multicolumn{1}{c}{\bf An. 5} & \multicolumn{1}{c}{\bf Avg.} \\\hline

\multicolumn{1}{c}{\textbf{Rule-based baseline}} & \multicolumn{1}{c}{0.748} & \multicolumn{1}{c}{0.716} & \multicolumn{1}{c}{0.661} & \multicolumn{1}{c}{0.715} & \multicolumn{1}{c}{0.723} & \multicolumn{1}{c}{0.713}\\

\multicolumn{1}{c}{\textbf{Supervised system}} & \multicolumn{1}{c}{0.984} & \multicolumn{1}{c}{0.905} & \multicolumn{1}{c}{0.906} & \multicolumn{1}{c}{0.938} & \multicolumn{1}{c}{0.945} & \multicolumn{1}{c}{0.936} \\

\multicolumn{1}{c}{\textbf{Proposed system}} & \multicolumn{1}{c}{0.991} & \multicolumn{1}{c}{0.970} & \multicolumn{1}{c}{0.975} & \multicolumn{1}{c}{0.982} & \multicolumn{1}{c}{0.990} & \multicolumn{1}{c}{0.982}\\\bottomrule
\end{tabular}
\end{table*}

\subsection{Application use case}

\begin{figure*}[!htbp]
\centering
\includegraphics[scale=0.2]{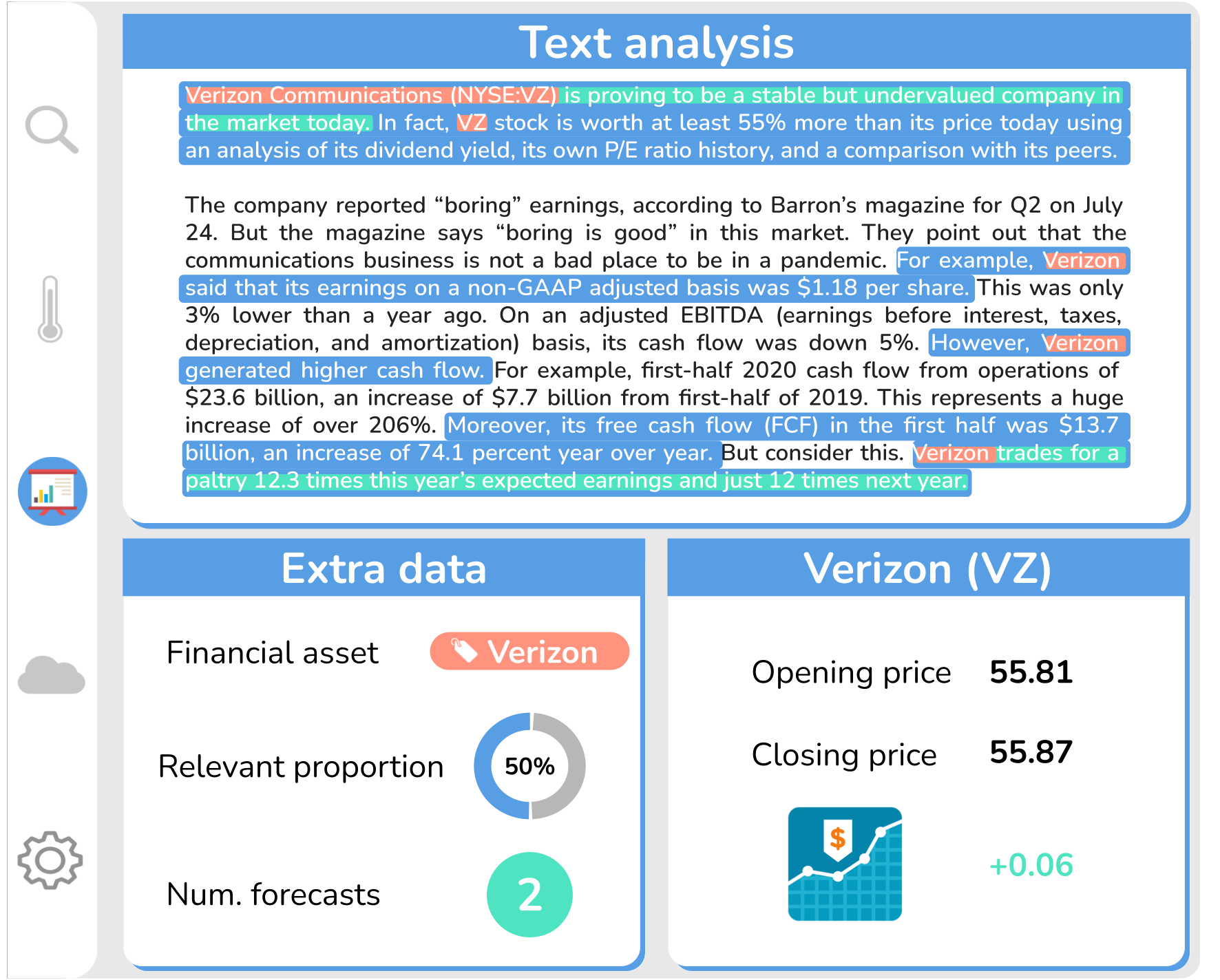}
\caption{\label{turnitin} Example of financial event detection.}
\end{figure*}

Figure \ref{turnitin} shows a news piece highlighted by our system. Relevant sentences are highlighted in blue, asset identifiers in pink and predictions/forecasts in green. Note the differences with the manual highlighting in Table \ref{news_use_case_annotated}. For example, human annotators might mark the sentence ``In fact, VZ stock is worth at least 55\% more than its price today'' as a prediction. With our system, however, both forecasts and relevant, informative sentences are marked.

The dashboard at the bottom of Figure 2 summarises the results, showing the proportion of relevant segments and number of predictions and forecasts. The updated value of the financial asset, taken from Yahoo Finance\footnote{Available at \url{https://finance.yahoo.com}, October 2022.}, is shown on the right.

\section{Conclusions}
\label{sec:conclusions}

Many valuable online financial news sources, such as economy journals and web pages (Motley Fool, InvestorDaily, etc.), contain opinions from experts describing relevant market events within sociological, political and/or cultural contexts. 

The system proposed in this paper is designed to extract this relevant information, and in particular forecasts and predictions. To do this, it employs \textsc{nlp} techniques. It segments the text and applies \textsc{lda} analysis to filter out less relevant sentences, and then applies discursive temporality analysis to identify predictions and forecasts within the remaining relevant text. The result is a summary of relevant, easy-to-read information. We are not aware of any other \textsc{ke} systems that have applied a similar approach to resolve this problem.

To our knowledge and considering related work, our proposal is the first to jointly consider relevance and temporality at the discursive level. It contributes to transferring human associative discourse capabilities to expert systems by combining (\textit{i}) multi-paragraph topic segmentation and co-reference resolution to separate author expression patterns, (\textit{ii}) detection of relevant text through topic modelling with \textsc{lda}, and (\textit{iii}) identification of forecasts and predictions within relevant text using discursive temporality analysis and \textsc{ml}.

We have created an experimental data set composed of 2,158 financial news items to evaluate our proposal. We have validated its annotation capacity by performing an inter-agreement analysis using Alpha-reliability and accuracy measures and evaluated its performance using the state-of-the-art \textsc{rouge} metric. The system attained \textsc{rouge-l} values of 0.662 and 0.982 for the detection of relevant data and predictions/forecasts, respectively. We also compared the performance of our system with a rule-based baseline system and a fully supervised system (which also performs supervised extraction of relevant text) to evaluate its competitiveness. It outperformed the rule-based system and was comparable to the fully supervised system, which unlike our solution requires manual annotation.

In future work, we plan to extend our research to Spanish and other languages to cover a broader community of investors. We will also evaluate the system in composite (multi-disciplinary) research domains.

\section*{Declarations}

\subsection*{Acknowledgement}
This version of the article has been accepted for publication, after peer review (when applicable) but is not the Version of Record and does not reflect post-acceptance improvements, or any corrections. The Version of Record is available online at: https://doi.org/10.1007/s10489-023-04452-4

\subsection*{Funding}

This work was partially supported by Xunta de Galicia grants ED481B-2021-118 and ED481B-2022-093, Spain. University of Vigo/CISUG covered the open access fee.

\subsection*{Competing Interests}
The authors have no relevant financial or non-financial interests to disclose.

\subsection*{Ethics approval}

Not applicable

\subsection*{Consent to participate}
Not applicable

\subsection*{Consent for publication}
Not applicable

\subsection*{Availability of data and material}

The data sets generated during and/or analysed during the current study are available from the corresponding author on reasonable request.

\subsubsection*{Code availability}

The code used in this work is not publicly available.

\subsubsection*{Authors' contributions}

\textbf{Silvia García-Méndez}: Conceptualisation, Methodology, Software, Validation, Formal analysis, Investigation, Data Curation, Writing - original draft. \textbf{Francisco de Arriba-Pérez}: Conceptualisation, Methodology, Software, Validation, Formal analysis, Investigation, Data Curation, Writing - original draft. \textbf{Ana Barros-Vila}: Software, Validation, Investigation, Resources, Data Curation, Writing - review \& editing. \textbf{Francisco J. González-Castaño}: Conceptualisation, Methodology, Data Curation, Writing - review \& editing, Supervision. \textbf{Enrique Costa-Montenegro}: Methodology, Data Curation, Writing - review \& editing.

\bibliography{bibliography}

\end{document}